%% file: main.tex
\documentclass{article}
\usepackage[final,sglblindworkshop]{neurips_2026}
\usepackage[utf8]{inputenc} 
\usepackage[T1]{fontenc}    
\usepackage{hyperref}       
\usepackage{url}            
\usepackage{booktabs}       
\usepackage{amsfonts}       
\usepackage{nicefrac}       
\usepackage{microtype}      
\usepackage{xcolor}         

\usepackage{comment}
\usepackage{amsmath} 
\usepackage{caption}
\usepackage{cleveref}
\usepackage{makecell}
\usepackage{graphicx}
\newcommand\blfootnote[1]{%
  \begin{NoHyper}%
  \renewcommand{\thefootnote}{}%
  \footnotetext{#1}%
  \end{NoHyper}%
}

\title{A Bag of Concepts? Testing Role-Filler Binding in the Verbalizable Workspace}
\workshoptitle{Interpretability as a Science}

\author{
  Reya Dawlah\thanks{Equal contribution.} \\
  University of Texas at Dallas \\
  \And
  Elizabeth McMillin\footnotemark[1] \\
  MiraCosta College \\
  \And
  Michael Shihong Zhang\footnotemark[1] \\
  University of Waterloo \\
  \AND
  Badrish Ananth \\
  University of California, Irvine \\
  \And
  Rishi Adi Ruia \\
  University of Washington \\
  \And
  Ane Zuñiga \\
  Massachusetts Institute of Technology \\
  \And
  Thomas Jiralerspong \\
  Mila, Universit\'e de Montr\'eal \\
}

\begin{document}

\maketitle
\blfootnote{Correspondence to \texttt{reya.dawlah@utdallas.edu} and \texttt{anezu@mit.edu}.}

\input{0_abstract/main}
\input{1_introduction/main}
\input{2_related_work/main}
\input{3_methodology/main}
\input{4_experimental_setup/main}
\input{5_results/main}
\input{6_discussion/main}
\input{7_conclusions/main}

\bibliographystyle{plainnat}
\bibliography{references}

\appendix
\input{appendix/main}

\end{document}

%% file: 0_abstract/main.tex
\begin{abstract}
Runtime alignment monitors built on a verbalizable workspace have been proposed as a way to surface a model's silent cognition before it produces output \citep{gurnee2026workspace}.
Their value rests on the workspace carrying enough structure to distinguish safety-relevant propositions, most of which are relational: \textit{the model plans to deceive the user} and \textit{the user plans to deceive the model} share a concept set and differ only in role assignment.
We ask whether the workspace carries the role-filler bindings (which entity fills which role) such a monitor would need, on three frontier language models (Qwen3.6-27B, Gemma-3-12B, Gemma-3-27B-IT), under two lens families (J-lens and R-lens), and with two methods for fitting the role direction (difference-of-means and LRE-gradient).
We find a clean dissociation: the workspace carries directions that steer what the model outputs about roles but not the representations the model uses to compute them.
Linear probes recover role at near-ceiling from the residual stream but score below a rank-matched random control when reading the workspace, and steering the model along the fitted role direction changes what it reports about roles but no more than strength-matched controls change which entity it binds to.
Ablating the workspace (zeroing the residual stream's projection onto its ten most active lens directions) damages binding on one of the three models (Gemma-3-27B-IT, holding across three lens variants) and shows no binding-specific damage on the other two, with a rank-matched random subspace as the control; this confirms our design detects binding when present, though scale and instruction tuning are confounded there.
A workspace-lens monitor built on these directions would therefore miss a role assignment the model itself represents.
\end{abstract}

%% file: 1_introduction/main.tex
\section{Introduction}
\label{sec:introduction}

Runtime alignment monitoring aims to surface a model's cognition during inference in a form a human can read, before the model produces any output. The verbalizable workspace has been proposed as a candidate substrate for this. \citet{gurnee2026workspace} identify a small subspace of the residual stream that carries information about the model's cognition in a form that can be projected into vocabulary space and read as natural language. If this holds, a monitor operating on that subspace during inference could surface strategic reasoning, evaluation awareness, or deceptive intent. \citet{Blank2026} have since introduced R-lens, which applies layer-wise relevance propagation to the backward pass to reduce error accumulation, yielding more faithful atoms on early layers and directions its authors report to be more causally loaded than J-lens ones.

We refer to this subspace as the workspace. Both lenses read the same residual stream at the same sites under the same workspace hypothesis, differing only in how the backward pass is computed; each returns its own top-$k$ atoms, and we write J-space and R-space for the bases they induce. We compare R-lens against the matched J-lens released alongside it, fitted on the same forward pass and corpus, so any R-versus-J difference is attributable to lens type rather than provenance.

The monitoring case rests on the workspace carrying enough structure to distinguish safety-relevant propositions from ones that are irrelevant to safety. Many such propositions are relational (about how entities relate to each other) rather than lexical (about which words are present): what makes them safety-relevant is not which concepts are present, but how those concepts are assigned to argument positions. The propositions \textit{the model plans to deceive the user} and \textit{the user plans to deceive the model} contain the same unordered set of concepts, $\{\text{model}, \text{user}, \text{plan}, \text{deceive}\}$. What distinguishes them is role-filler binding, the assignment of entities to their positions in a relational structure, in this case which entity is the agent and which is the patient of the verb \citep{greff2020binding, feng2024language}. A monitor that reads co-active concepts without their bindings cannot tell the two propositions apart. Whether the workspace carries role-filler binding, or only a bag of concepts, was left as an open question by \citet{gurnee2026workspace}.

\begin{figure}[t]
    \centering
    \includegraphics[width=\linewidth]{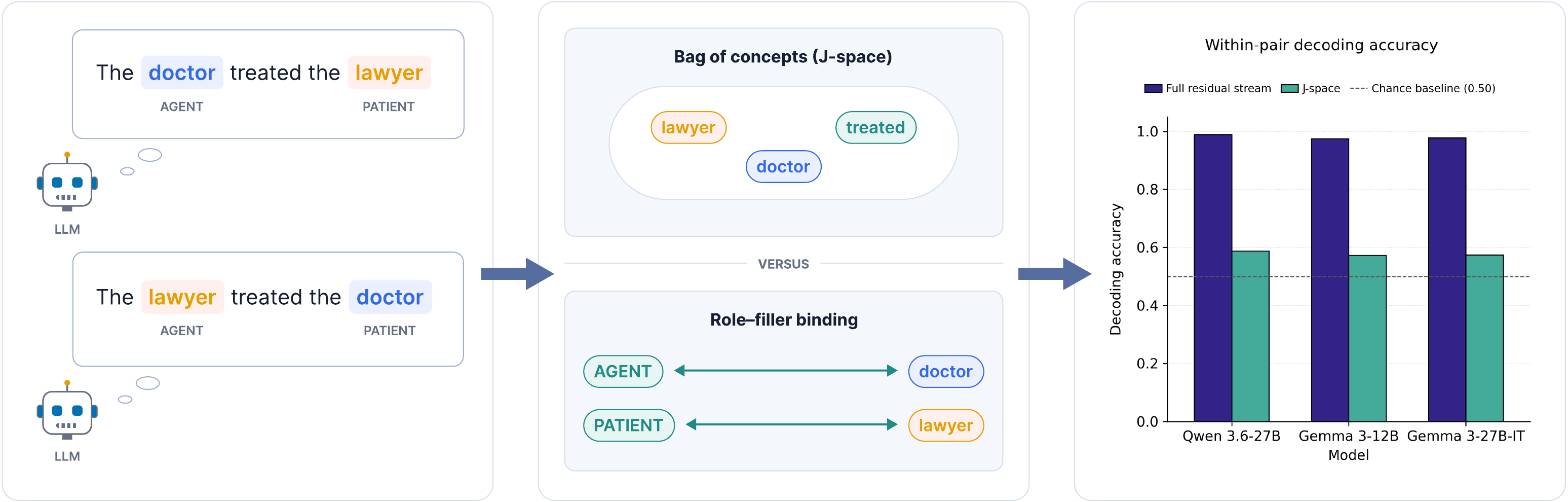}
    \caption{\textit{(Left)} Two sentences share a concept set but assign roles oppositely. \textit{(Center)} A bag-of-concepts workspace preserves the concepts but loses the role assignment; a role-filler binding workspace keeps both. \textit{(Right)} We show that a linear probe recovers role at near-ceiling accuracy from the residual stream and close to chance from J-space, for all three models.}
    \label{fig:teaser}
\end{figure}

We evaluate role-filler binding inside the workspace on three frontier language models (\texttt{Qwen3.6-27B}, \texttt{Gemma-3-12B}, and \texttt{Gemma-3-27B-IT}), under two projection tools (J-lens and R-lens), and with two estimators for the relational direction: difference of means, and a readout-gradient estimator in the style of Linear Relational Embeddings \citep{hernandez2024linearity}, following the concept-direction adaptation of \citet{chanin2024identifying}. The main finding is a dissociation between readout and computation. Readout refers to what the model reports about roles, measured by whether pushing a fitted role direction in the workspace shifts which entity it names as agent. Computation refers to the representation the model uses to assign those roles in the first place, measured by whether role is linearly decodable from the workspace and whether removing the workspace damages binding. The workspace supports readout on two of the three models, while carrying less role information than a rank-matched random subspace on all three. \Cref{fig:teaser} shows the question and previews the answer.

Our main contributions are:
\begin{itemize}
\item A causal null on role-filler binding under intervention in the workspace, alongside a narrow but real signed steering effect along the same directions at the same sites on two of three models, dissociating what the model outputs about roles from what it uses to assign them.
\item A causal evaluation design that separates a real absence of binding structure in the workspace from three alternative explanations, a weak estimator, a tool-specific artifact, and a model-specific artifact, with internal positive controls that fire when a causal effect is present.
\item Evidence that a workspace lens monitor surfaces the model's role reports rather than its bindings, bounding the runtime alignment monitoring case for the workspace lens paradigm on safety-relevant propositions whose benign and problematic states share a concept set.
\end{itemize}

\noindent Code, data, and analysis are available at \url{https://github.com/Reyadawlah/role-filler-binding-jspace}.

%% file: 2_related_work/main.tex
\section{Related Work}
\label{sec:related_work}

\input{2_related_work/1_j_space}
\input{2_related_work/2_r_lens}
\input{2_related_work/3_binding}
\input{2_related_work/4_estimators}

%% file: 2_related_work/1_j_space.tex
\paragraph{The J-lens and the workspace}
\label{sec:related_work:jlens}

Interpreting what intermediate layers in a language model represent has progressed through multiple stages. The logit lens \citep{nostalgebraist2020} projects intermediate activations through the unembedding matrix, giving us insight into which tokens the model tends towards at each layer. The tuned lens \citep{belrose2025elicitinglatentpredictionstransformers} learns an affine transformation that maps intermediate layer activations onto the final-layer coordinate system without requiring a full Jacobian computation. Sparse autoencoders \citep{bricken2023monosemanticity} decompose activations into features that are more selective and interpretable to humans, with the cost of manual feature identification per concept. The J-lens \citep{gurnee2026workspace} narrows this further by identifying a small verbalizable subspace of the model's residual stream called the J-space. The J-lens associates each vocabulary token with the direction in the residual stream whose activation on average causally increases that token's logits. This direction is computed by taking an average Jacobian of the final-layer hidden state with respect to intermediate activations, then multiplying by the unembedding matrix. The J-space accounts for 6--7\% of activation variance but is causally decisive on concept-swap interventions.

The J-lens readout is a set rather than a structure: variations of the same sentence output approximately the same top concepts, without any way to distinguish between who performed the action and who received it. \citet{gurnee2026workspace} acknowledge this explicitly and leave open the question of whether the concepts in the workspace are \emph{bound} into semantic roles, and we aim to test that question directly.

%% file: 2_related_work/2_r_lens.tex
\paragraph{The R-lens}
\label{sec:related_work:rlens}

Where the J-lens reads the workspace by projecting activations onto vocabulary-indexed directions via the average Jacobian, the R-lens \citep{Blank2026} is a structural extension of the same paradigm. Rather than indexing the workspace directions by the vocabulary tokens, it uses layer-wise relevance propagation to index them by relation. In other words, the J-lens identifies which concepts are active and the R-lens asks which relational structures exist.

%% file: 2_related_work/3_binding.tex
\paragraph{Binding in language models}
\label{sec:related_work:binding}
Several works investigate role-filler binding in language models. \cite{feng2024language} introduce binding IDs and demonstrate that role filler information is recoverable from the full residual stream. \cite{feng2025monitoring} extend this further by introducing propositional probes, representing the closest prior work to ours. Their work identifies subspaces that contain propositional information from model activations post hoc, whereas we ask whether binding is already represented in a pre-existing workspace, the J-space. Our experiment tests binding specifically within this verbalizable workspace instead of discovering a binding subspace from the residual stream. \cite{greff2020binding} provide theoretical grounding to distinguish between bag-of-concept representation and relational representations, and motivate our tests of whether concepts in the J-space are bound to semantic roles.

%% file: 2_related_work/4_estimators.tex
\paragraph{Estimators for causal directions}
\label{sec:related_work:estimators}
How interpretable directions are extracted affects the conclusions of causal experiments. Standard probing methods learn a classifier that separates examples based on their activations, but probe accuracy alone can reflect probe expressivity rather than the structure of the representation \citep{hewitt2019designing, pimentel2020information}. \citet{chanin2024identifying} introduce Linear Relational Concepts (LRC) and Linear Relational Embeddings (LRE), which model relationships between representations and use those relationships to construct interpretable concept directions. Their method provides a more principled alternative to taking simple differences between class means.

%% file: 3_methodology/main.tex
\section{Methodology}
\label{sec:method}

We distinguish two things a model can do with role information. \textbf{Readout} is what the model outputs about roles, i.e., when the model is asked \textit{who did the treating}, it answers \textit{the doctor}. \textbf{Computation} is what the model uses internally to assign the entities to their roles in the first place. The workspace can carry directions that steer the readout without hosting the substrate the model uses to compute the binding, and our design tests both.
The design has three experiments: a linear-probe test for decodability, an intervention test with a binding-score metric, and an ablation test for causal necessity.
\input{3_methodology/1_stimuli}
\input{3_methodology/2_probes}
\input{3_methodology/3_fit}
\input{3_methodology/4_intervention}
\input{3_methodology/5_binding}
\input{3_methodology/6_stats}
\input{3_methodology/7_ablation}

%% file: 3_methodology/1_stimuli.tex
\paragraph{Stimulus set and task}
\label{sec:method:stimuli}
We constructed a stimulus set where each item is a family of four minimal-pair sentences crossing role (agent, patient) with linear position (first, second) of a target entity, across four syntactic constructions: active or passive, cleft, relative clause, and dative. Because each family treats the entities as fixed while varying both their position in the sentence and the construction of the sentence as a whole, averaging over position within a family cancels word-order effects and averaging over construction removes any effects from active/passive voice or syntactic framing, so role assignment is the only factor which remains once both have been taken out of the picture. The direction-fitting corpus is disjoint from the evaluation corpus at both the template and entity level, so no direction is ever fit and evaluated on the same sentence.

Given a stimulus sentence, the model is prompted with a role probe and returns the next-token distribution (for example, \textit{Question: Who treated someone? Answer: The doctor}). We read the probability of the target entity token and its partner from the softmax over the full vocabulary, and convert to log-odds. Every subsequent measurement is built upon these two probabilities.

%% file: 3_methodology/2_probes.tex
\paragraph{Reading the workspace.}
\label{sec:method:probes}
We first ask whether role information is linearly decodable from the workspace at all. At the read layer, we extract four activation sources from the residual-stream activation  at either the entity token or the sentence-final token:
\begin{itemize}
    \item \textbf{Full residual stream.} This is essentially the raw activation $h$.
    \item \textbf{J-space.} The sparse nonnegative J-space component of $h$, recovered by orthogonal matching pursuit \citep{pati1993orthogonal} against the J-lens dictionary. This approximates the gradient pursuit of \cite{gurnee2026workspace}. See \Cref{app:matching_pursuit} for the algorithm.
    \item \textbf{Orthogonal remainder.} $h$ minus its J-space component, which translates to everything the lens cannot see.
    \item \textbf{Matched-rank random subspace.} $h$ projected onto a seeded random orthonormal basis of matched rank. Since the J-space has a small effective rank whereas the residual has the full model dimension, a direct comparison would confound content with capacity.
\end{itemize}
The random subspace basis is drawn once per run rather than per stimulus. It is a readout basis, like J-space itself, and treating it that way keeps the comparison fair.

The probe is closed-form ridge regression to $\pm 1$ role labels, with a bias term. Cross-validation is leave-one-pair-out, meaning that each fold holds out all stimuli of one concept pair, so above-chance (greater than 0.5) held-out accuracy requires role information that generalizes across lexical items.

We also run two robustness checks to eliminate the possibility that a weak probe is responsible for any negative findings. The first varies the probe class, since tensor-product binding \citep{smolensky1990tensor} represents a role-filler pair as the product of a role vector and a filler vector, and reading the role back requires a decision function that multiplies activation dimensions instead of weighting each one independently. Our first check uses a quadratic probe, which can form that product, whereas a linear one cannot. We also add a random-Fourier-features probe to cover nonlinear structure which the quadratic probe would miss. With our second check, we vary regularization by testing seven different values for the ridge penalty under nested cross-validation. This ensures that a negative result cannot be attributed to a badly chosen penalty.

%% file: 3_methodology/3_fit.tex
\paragraph{Fitting the role direction}
\label{sec:method:fit}
The intervention nudges activations along a fitted role direction $r$. We fit $r$ per entity per site using two estimators, since running both lets us distinguish a null result about the workspace from a null result about a weak estimator.

The difference-of-means estimator subtracts the mean activation under the patient role from the mean under the agent role, over the fitting corpus.
This finds the direction separating the two role classes.

The LRE gradient estimator, adapted from \cite{chanin2024identifying}, fits the direction the readout actually responds to rather than the direction that separates classes. For each exemplar we take the gradient of the readout $\partial(z_{\text{entity}} - z_{\text{other}})/\partial h$ at the site token, where $z_{\text{entity}}$ and $z_{\text{other}}$ are the logits of the two answer tokens under the role probe. We pool exemplars across both roles (the contrast lives inside each gradient, so no class subtraction is needed) and take the unit mean:
\begin{equation}
r^{\text{LRE}}_{\text{entity}} = \mathrm{unit}\!\left(\frac{1}{n}\sum_{i=1}^{n} \nabla_h (z_{\text{entity}} - z_{\text{other}})\big|_{\text{exemplar } i}\right).
\end{equation}

Both estimators produce unit vectors, so a push of magnitude $c$ perturbs the residual stream by exactly $c$ in norm, and every strength-matched control is norm-matched by construction. We also check that each fitted direction is stable by ensuring that the mean cosine between bootstrap-resampled fits and the full fit clears a pre-registered threshold before we use the direction.

%% file: 3_methodology/4_intervention.tex
\paragraph{Intervention}
\label{sec:method:intervention}
At the read-layer residual stream, at the injection site's token position, the push edit adds $h \leftarrow h + s \cdot c \cdot r_{\text{entity}}$, where $r_{\text{entity}}$ is the unit-vector direction, $s \in \{+1, -1\}$ selects toward-agent or toward-patient, and $c$ is the calibrated push coefficient. Every family receives both a positive and a negative push. The same $(r, c)$ applies to each cell in the family, never conditioned on the sentence's own role label. This is why a role-dependent effect can be attributed to binding.

We compare the real push (\textsc{role\_push}) against three strength-matched controls that share the same $(c, s)$ machinery and differ only in which direction is pushed:

\begin{itemize}
    \item \textbf{Non-participant push.} The fitted direction of an entity that does not appear in the sentence. Same estimator, fit, and magnitude, but the pushed concept has no role to bind.
    \item \textbf{Random direction.} A seeded random unit vector at the same magnitude. This rules out the chance that any push of that size, at that site, produces the same effect.
    \item \textbf{Shuffled-label direction.} For diff-of-means, this is the direction refit after permuting the agent and patient labels. For LRE, this is the pooled gradient with signs scrambled at random (label permutation is inert to a pooled mean). The purpose of this is to eliminate the possibility that the fitted direction captures corpus quirks rather than role.
\end{itemize}

A no-edit baseline is also collected that provides each family's natural log-odds gap between the two role sentences, which the binding score subtracts.

A separate identity-swap edit patches in lens coordinates by swapping the J-lens components of two tokens (\emph{doctor} into \emph{nurse}). Note that this is not a role intervention and that we run it as an intervention-strength check on the editing machinery, following the source paper's own use of it. We treat its outcome separately from the push results.

\paragraph{Calibrating the push coefficient.} We select a $c$ value per model, estimator, and site on the fitting corpus, disjoint from the evaluation stimuli. The rule is the smallest value on a doubling grid whose toward-agent push shifts the entity's log-odds on patient-role fitting sentences by at least a pre-registered minimum. The floor rules out interventions too weak to move behavior, so a null binding score on an intervention that fails to clear the floor is uninterpretable in either direction.

\paragraph{Strength check.} At analysis time we re-verify strength of the intervention on the evaluation stimuli. The mean absolute log-odds displacement of the readout under \textsc{role\_push} must clear the pre-registered floor, and a bootstrap CI on the paired difference against the strongest control must exclude zero. Passing this check authorizes the binding claim, while failing it produces the \emph{uninterpretable strength failure} verdict described below.

%% file: 3_methodology/5_binding.tex
\paragraph{Binding score}
\label{sec:method:score}
The binding score measures the role-dependent, edit-induced shift in the log-odds gap between the two role sentences of a family. It subtracts the natural gap that exists before any intervention, so what survives is only the change the edit introduces.

Let $L(\text{role}, \text{edit}, \text{sign})$ denote the log-odds of the target entity under the role probe, averaged over the two surface positions of the family. Position averaging cancels surface-order bias, so a role effect that survives it cannot be a word-order artifact. The per-sign gap change and the family binding score are
\begin{align}
dG(s) &= \left[L(\text{agent}, \text{push}_s) - L(\text{patient}, \text{push}_s)\right] \nonumber - \left[L(\text{agent}, \text{no edit}) - L(\text{patient}, \text{no edit})\right], \\
\text{BS}_i &= -\left(dG(\text{toward agent}) + dG(\text{toward patient})\right) / 2.
\end{align}

A workspace that only holds a bag of concepts leaves $dG(s)$ near zero for both signs. A workspace that binds shrinks the natural gap from opposite sides when pushed in opposite directions, making both signed gap changes negative and $\text{BS}_i$ positive. The two-sign design also surfaces asymmetries, if promotion and demotion move the score differently. Appendix~\ref{app:worked-example} walks through one stimulus family end to end, showing the probe, the four cells' log-odds, and how the sign convention above lands on real numbers.

The same statistic, with \textsc{role\_push} replaced by each control edit in turn, produces the null distribution. The two-tailed percentile band of the control scores is the \textbf{null band}. A binding score is credible only if the pooled mean clears the band; clearing zero is not enough.

%% file: 3_methodology/6_stats.tex
\paragraph{Statistics}
\label{sec:method:stats}
The unit of resampling and permutation is the stimulus family: each family contributes one $\text{BS}_i$, the difference-in-differences comparison is computed within the score, and family-level resampling preserves it. We report bootstrap CIs on the pooled mean and on Cohen's $d$ against zero. The primary significance test is a two-sided sign-flip permutation test; within a family, agent and patient labels are exchangeable, so a sign flip on $\text{BS}_i$ is an exact draw from the null. Holm-Bonferroni correction is applied over per-construction and per-(construction, pair) p-values.

Rather than reporting a binary yes/no on significance, the pooled result is classified into one of four outcomes, dependent on the push arm's strength check:

\begin{itemize}
    \item \textbf{Positive binding.} Requires a significant pooled test, a meaningful effect size, a pooled mean above the null band, the effect being consistent across at least two of the three agent/patient constructions, and a random-direction control mean within the band.
    \item \textbf{Indistinguishable from controls.} Involves a significant pooled test, but one where the pooled mean sits inside the null band. Cohen's $d$ here measures how consistently the push moves the score, not whether the movement has anything to do with role.
    \item \textbf{Clean negative.} Not significant, with a tight CI on Cohen's $d$. Acts as evidence for the bag-of-concepts answer, not just absence of evidence for binding.
    \item \textbf{Uninterpretable strength failure.} The role-push arm did not move the readout above the pre-registered floor or did not beat the strongest control. No binding conclusion follows.
\end{itemize}

%% file: 3_methodology/7_ablation.tex
\paragraph{Ablation}
\label{sec:method:ablation}
We also test whether the workspace is causally load-bearing for binding by ablating it and measuring the effect on the model's own binding performance. During the forward pass at the read layer, we zero the projection of the residual stream onto the span of the most active J-lens vectors, chosen per stimulus by the same matching-pursuit procedure used to define J-space. We score two tasks in log-odds: binding (role probe, signed agent-versus-patient gap) and recall (concept probe, asking which participant matches a professional cue, counterbalanced across both participants to eliminate base-rate confounds).

Let $M(\text{task}, \text{edit})$ denote the mean log-odds margin on a task under an edit condition. The binding-specific deficit for ablation $a$ is
\begin{equation}
\Delta_a = \frac{M(\text{binding}, \text{no edit}) - M(\text{binding}, a)}{M(\text{binding}, \text{no edit})}
       - \frac{M(\text{recall}, \text{no edit}) - M(\text{recall}, a)}{M(\text{recall}, \text{no edit})}.
\end{equation}
Each term is the drop in that task's margin as a share of its own no-edit baseline. Normalizing keeps task magnitudes from distorting the subtraction. The matched control is a random orthonormal subspace ablation of the same rank, drawn fresh per family with a seeded RNG. Causal involvement requires $\Delta_{\text{J-space}}$ to exceed a pre-registered magnitude floor, exceed $\Delta_{\text{random}}$, and have a bootstrap CI that excludes zero. Baseline accuracy on binding and recall must also be within a pre-registered tolerance, so that a large binding-specific deficit cannot be attributed to the recall task being easier.

%% file: 4_experimental_setup/main.tex
\section{Experimental Setup}
\label{sec:experimental_setup}

\input{4_experimental_setup/1_models}
\input{4_experimental_setup/2_layer}
\input{4_experimental_setup/3_stimuli}
\input{4_experimental_setup/4_probes}
\input{4_experimental_setup/5_intervention}
\input{4_experimental_setup/6_analysis}

%% file: 4_experimental_setup/1_models.tex
\paragraph{Models, tokenizers, and lenses}
\label{sec:setup:models}
We ran experiments on three dense language models: Qwen3.6-27B \citep{qwen3.6-27b} (\texttt{Qwen/Qwen3.6-27B}, 64 layers, $d_{\text{model}} = 5120$) and two Gemma-3 variants \citep{team2025gemma}: Gemma-3-12B (\texttt{google/gemma-3-12b-pt}, 48 layers) and Gemma-3-27B-IT (\texttt{google/gemma-3-27b-it}, 62 layers). Qwen3.6-27B and Gemma-3-12B are base models; Gemma-3-27B-IT is the only instruction-tuned model in the set. \cite{Blank2026} confirm that the J-lens phenomenon replicates on Qwen3.6-27B, one of our motivations for including it. Every entity token used in stimuli was verified to tokenize as a single token in both the Qwen and Gemma tokenizers.

For J-lens readouts on Qwen3.6-27B and Gemma-3-12B we used pre-fitted per-layer Jacobians \citep{neuronpedia_jacobian_lens}, both fitted on WikiText-103 \citep{merity2016pointer} with 1000 prompts. For Gemma-3-27B-IT we ran three lens variants (the J-lens from the same release, and an R-lens and matched J-lens from \cite{Blank2026}).

%% file: 4_experimental_setup/2_layer.tex
\paragraph{Read layer and injection site}
\label{sec:setup:layer}

We picked the primary read layer per model to match the single-layer analyses in the source paper \citep{gurnee2026workspace}, which sit at approximately 75\% relative depth. Each primary experiment is single-layer, applying intervention edits and the ablation only at the read layer. For the layer sweep on Gemma-3-27B-IT, we additionally evaluated the workspace band edges in line with the source paper \citep{gurnee2026workspace}. Specific layer indices and rounding conventions are in Appendix~\ref{app:hyperparameters}.

Interventions and probes targeted one of two token positions per sentence: the entity token (last subtoken of the target entity word) or the sentence-final token. Primary results use the entity token, and comparisons to the sentence-final site are reported only where informative.

%% file: 4_experimental_setup/3_stimuli.tex
\paragraph{Stimulus set}
\label{sec:setup:stimuli}
The primary stimulus set uses three concept pairs: \texttt{doctor}/\texttt{nurse}, \texttt{teacher}/\texttt{student}, \texttt{driver}/\texttt{passenger}. The design crosses these with four constructions (active-passive, cleft, relative clause, dative) and the 2$\times$2 role$\times$position cells, giving 50 families per (pair, construction) cell, for 600 families and 2400 sentences per model.

The primary Qwen cross-pair transfer results, including the entity-token full-residual accuracy of 0.594 (Appendix~\ref{app:crosspair}), use the three-pair stimulus set above. A separate Qwen robustness run uses six pairs, adding \texttt{lawyer}/\texttt{judge}, \texttt{pilot}/\texttt{nurse}, and \texttt{coach}/\texttt{student}. To keep the total item budget constant, we used 25 families per cell instead of 50. The non-participant push control uses \texttt{chef}, \texttt{farmer}, and \texttt{coach} in the three-pair runs and \texttt{chef} and \texttt{farmer} in the six-pair run.

The direction-fitting corpus is disjoint from the evaluation corpus at both template and entity level.

%% file: 4_experimental_setup/4_probes.tex
\paragraph{Probing and direction fitting}
\label{sec:setup:probes_fit}

The J-space slice keeps the top $K = 16$ J-lens atoms, and the matched-rank random subspace has the same rank. The ridge probe uses $l_2 = 10^{-2}$; nonlinear probes cover quadratic and random-Fourier variants. Direction fitting used 24 exemplars per role for the Qwen difference-of-means run and 48 for all LRE runs. Directions below a bootstrap stability threshold on mean cosine are not used. Full hyperparameters are in Appendix~\ref{app:hyperparameters}.

%% file: 4_experimental_setup/5_intervention.tex
\paragraph{Intervention, ablation, and calibration}
\label{sec:setup:intervention}
Push coefficients were calibrated per (model, estimator, site) on the fitting corpus against a pre-registered strength floor of 0.5 log-odds. Calibration for the identity-swap strength failed on all three models (counterpart-probability shifts of 0.0001 to 0.0048, below the 0.05 floor), so those runs use a full swap rather than a calibrated scaled version. Ablation uses the top J-lens vectors selected per stimulus, against a random-subspace control of matched rank drawn fresh per family. Calibration details are in Appendix~\ref{app:hyperparameters}.

%% file: 4_experimental_setup/6_analysis.tex
\paragraph{Analysis and compute}
\label{sec:setup:analysis}

Statistical tests use bootstrap resampling and sign-flip permutation on family-level binding scores, with Holm-Bonferroni \citep{holm1979simple} correction across (construction, pair) cells for the group-level table and across constructions for the per-construction standard. Because J-space, orthogonal, and residual accuracies are deterministic once the layer is fixed, the seed varies only the random-subspace draw, and we report its mean and range across seeds. The verdict classifier evaluates outcomes against pre-registered thresholds on effect size, CI width, and cross-construction consistency. Analysis and compute details are in Appendix~\ref{app:hyperparameters}.

%% file: 5_results/main.tex
\section{Results}
\label{sec:results}

\input{5_results/1_decodability}
\input{5_results/2_intervention}
\input{5_results/3_ablation}

%% file: 5_results/1_decodability.tex
\subsection{Role information sits outside J-space}
\label{sec:results:decodability}

A linear probe recovers role at near-ceiling accuracy from the residual stream, at nearly the same accuracy from the part of the stream the Jacobian lens discards, and at chance from the part the lens selects. \Cref{tab:dissociation} shows the pattern at the sentence-final position on all three models, where J-space has lower accuracy than the stream it originates from while performing at or below a random subspace of equal rank (model spread = 0.014).

\input{tables/tab_dissociation}

The J-space null is not a probe-capacity artifact. Using nonlinear probes and ridge-penalty sweeps do not lift J-space above the matched random subspace, and the pattern holds across seeds and layers within the workspace band (\Cref{app:robustness}). Because the sentence-final leave-one-pair-out cross-pair transfer inverts below chance for all activation sources, the inversion reflects the probing setup rather than J-space itself (\Cref{app:crosspair}).

%% file: tables/tab_dissociation.tex
\begin{table}[htb]
\captionsetup{labelfont=bf,textfont=it}
\caption{Within-pair role decoding accuracy at the sentence-final position. Chance is 0.50.}
\label{tab:dissociation}
\centering
\vspace{6pt}
\small
\begin{tabular}{lccc}
\toprule
Decoded from & Qwen3.6-27B & Gemma-3-12B & Gemma-3-27B-IT \\
\midrule
Full residual stream            & 0.989 & 0.975 & 0.978 \\
Orthogonal remainder            & 0.984 & 0.949 & 0.974 \\
Random subspace (matched rank)  & 0.643 & 0.589 & 0.590 \\
J-space                         & 0.587 & 0.573 & 0.574 \\
\bottomrule
\end{tabular}
\end{table}

%% file: 5_results/2_intervention.tex
\subsection{Steering the readout does not move binding}
\label{sec:results:intervention}

Under the LRE gradient estimator, direction stability, measured as mean cosine similarity, ran from 0.967 to 0.999 at both sites in all three models. A three-way comparison panel on Gemma-3-12B moves the readout by $+0.94$ log-odds under agent-push, $-0.99$ under patient-push, and $-0.01$ under a matched random direction. Under the difference-of-means estimator, direction stability at the sentence-final site ran from 0.580 to 0.679, below our 0.8 threshold, and the comparison panel showed no readout displacement.

\input{tables/tab_steering}

\Cref{tab:steering} shows the LRE coefficient sweep on Gemma-3-27B-IT at the entity token. The pooled effect grows more than tenfold across the sweep (as $c$ goes from 64 to 256), but the non-participant control grows with it: the ratio of real-push to control stays near one across the sweep, and the pooled mean stays inside the null band throughout.
The other two models give the same pattern under LRE. On Qwen3.6-27B at $c = 4$, the pooled effect is $-0.0006$ and the non-participant control is $-0.0046$, without the signed steering seen on the other two models. On Gemma-3-12B at $c = 256$, the pooled effect is $-0.0140$ and the control is $-0.0308$. The verdict classifier returns \emph{indistinguishable from controls} on all three models.
The identity-swap calibration returned counterpart-probability shifts of 0.0012 on Qwen3.6-27B, 0.0048 on Gemma-3-12B, and 0.0001 on Gemma-3-27B-IT, all below the 0.05 floor.

%% file: tables/tab_steering.tex
\begin{table}[htb]
\captionsetup{labelfont=bf,textfont=it}
\caption{LRE push coefficient sweep on Gemma-3-27B-IT at the entity token. Ratio is pooled-mean over non-participant control.}
\label{tab:steering}
\centering
\vspace{6pt}
\small
\begin{tabular}{ccccc}
\toprule
$c$ & Displacement & Pooled mean & Non-participant control & Ratio \\
\midrule
64  & 0.328         & $-0.0134$ & $-0.0161$ & 0.83 \\
128 & 0.645         & $-0.0568$ & $-0.0561$ & 1.01 \\
256 & 1.255         & $-0.1809$ & $-0.2161$ & 0.84 \\
\bottomrule
\end{tabular}
\end{table}

%% file: 5_results/3_ablation.tex
\subsection{Ablating J-space damages binding in one model of three}
\label{sec:results:ablation}

We zeroed the top-10 J-space directions during the forward pass and measured the binding-specific deficit relative to a matched-rank random-subspace ablation. \Cref{tab:ablation} reports both sites on all models.

\input{tables/tab_ablation}

Qwen3.6-27B and Gemma-3-12B show no binding-specific damage at either site. The Qwen J-space edit shifted the residual norm around 5 times harder than the random edit did without additional binding damage. On Gemma-3-12B at the entity token, concept accuracy fell along with role accuracy, which resulted in the recall control failing the difficulty-matching precondition.

Gemma-3-27B-IT at the entity token gives the exception: a binding-specific deficit of $+0.275$, 95\% CI $[0.262, 0.288]$, with concept accuracy holding at 0.920 and the two edits' residual-norm perturbations within 7\% of each other. At the sentence-final site, the deficit is $-0.001$. The asymmetry replicates across three lens variants (binding deficits 0.284, 0.334, 0.297; Appendix~\ref{app:rlens}).

%% file: tables/tab_ablation.tex
\begin{table}[htb]
\centering
\captionsetup{labelfont=bf,textfont=it}
\caption{Binding-specific deficit at the entity and sentence-final tokens, for Qwen3.6-27B, Gemma-3-12B, and Gemma-3-27B-IT. Rows marked $^{\dagger}$ fail the difficulty-matching precondition (binding and recall baseline accuracies within 0.02 of each other).}
\label{tab:ablation}
\small
\begin{tabular}{llccccc}
\toprule
Model & Site & \makecell{Role acc.\\no edit} & \makecell{Role acc.\\abl.\ J} & \makecell{Deficit\\(J)} & \makecell{Same,\\random} & \makecell{Edit mag.\\ratio} \\
\midrule
Qwen3.6-27B    & entity & 0.790 & 0.768 & $+0.014$ & $-0.009$ & 5.39 \\
Qwen3.6-27B $^{\dagger}$    & final  & 0.790 & 0.792 & $+0.000$ & $-0.007$ & 4.97 \\
Gemma-3-12B    & final  & 0.883 & 0.883 & $+0.008$ & $+0.001$ & 0.79 \\
Gemma-3-12B $^{\dagger}$    & entity & 0.883 & 0.736 & $+0.273$ & $+0.101$ & 0.94 \\
Gemma-3-27B-IT & final  & 0.990 & 0.991 & $-0.001$ & $-0.000$ & 0.88 \\
Gemma-3-27B-IT & entity & 0.990 & 0.740 & $+0.275$ & $+0.006$ & 0.94 \\
\bottomrule
\end{tabular}
\end{table}

%% file: 6_discussion/main.tex
\section{Discussion}
\label{sec:discussion}

\paragraph{Readout versus computation.}
Across three models, two lenses, and two direction estimators, the workspace carries directions that steer what the model outputs about agent-patient role assignments but does not carry the substrate the model uses to assign them. The residual stream decodes role at near-ceiling accuracy, and so does the orthogonal remainder that the Jacobian lens discards; the J-space slice decodes at or below a matched-rank random subspace. Nonlinear probes, ridge penalty sweeps, and seed and layer variations do not close the gap. Under the LRE gradient estimator the fitted role direction moves the model's output about roles in a signed and reversible way, but pushing along that direction does not move the binding score beyond what a non-participant control produces. The decoding side of this dissociation does not depend on an editing intervention, but the steering side does, and the identity-swap check used to validate the editing machinery fails on all three models, so the steering null is consistent with the LRE direction being irrelevant to binding and with the edit being too weak to reach it.

\paragraph{The one positive causal result.}
Ablating J-space on Gemma-3-27B-IT at the entity token damages binding specifically, and the effect replicates across three lens variants. This is the only causal signal in the study, and it fits inside the frame rather than contradicting it: the workspace can be causally load-bearing for binding on some models even when the direction fitted to it does not steer binding on any. What the positive result establishes is that the ablation design detects binding involvement when it is present, so the nulls on the other two models are genuine absences of effect rather than artifacts of a weak intervention. The result is confounded between instruction-tuning and model scale, and we do not resolve which one the workspace-load-bearing signal is a property of.

\paragraph{Implication for workspace-lens monitoring.}
Runtime alignment monitors built on workspace-lens readouts have been proposed as a way to surface silent cognition before the model produces output. Our results bound what such a monitor can catch on the class of relational propositions motivating it. Most safety-relevant propositions are relational: \textit{the model plans to deceive the user} and \textit{the user plans to deceive the model} contain the same set of concepts and differ only in role assignment. A monitor that reads workspace directions surfaces the concepts and the direction the readout is moving in; it does not surface the binding. On propositions where benign and problematic states share a concept set, such a monitor would respond similarly to both. The workspace-lens paradigm carries useful information about what the model is producing, and less than it appears to about what the model is computing.

\paragraph{Position against prior work.}
The closest prior work is \citet{feng2025monitoring}, which discovers a binding subspace from the residual stream post hoc, whereas we test whether binding is already represented inside a pre-existing, independently identified workspace, and find that it is not. \citet{greff2020binding} argue on theoretical grounds that distributed, superposed representations struggle to preserve role-filler independence; our null is what their account would predict for a small verbalizable subspace expected to summarize the model's cognition into a bag of concepts.

\paragraph{Limitations.}
The identity-swap intervention-strength check failed on all three models, which limits how far the steering null can be pushed. A follow-up positive control would patch in the residual from the opposite-role sentence at the same site and calibrate the push against that, which would give the steering arm an independently-grounded target. The positive ablation on Gemma-3-27B-IT is confounded between instruction-tuning and model scale; a base (non-instruction-tuned) Gemma-3-27B ablation would resolve this. All three models are dense transformers on English agent-patient stimuli at a single layer per model, and whether the dissociation holds on other architectures, layers, languages, or richer relational structures such as coreference, temporal relations, or multi-entity planning is untested. A comparison against Natural Language Autoencoders \citep{FraserTaliente2026} on Gemma-3-27B-IT would test whether the dissociation is specific to the workspace-lens paradigm or general to verbalizable readouts. 

%% file: 7_conclusions/main.tex
\section{Conclusion}
\label{sec:conclusion}

We asked whether the verbalizable workspace carries role-filler bindings, which a runtime alignment monitor would need to distinguish safety-relevant propositions that share a concept set.
To answer this, we ran decodability, intervention, and ablation tests on three frontier models, under two lenses and two direction estimators.
We find that the workspace carries a readout of role but not the substrate the model uses to compute it.
Role is decodable from the residual stream at near-ceiling accuracy, while from the workspace it is decodable below a rank-matched random control.
Directions fitted inside the workspace steer what the model reports about roles, but we detect no effect on which entity it binds as agent that strength-matched controls do not also elicit.
On the agent-patient bindings we test, a monitor built on workspace directions would therefore surface the model’s role reports rather than its bindings; whether the same holds on realistic monitoring scenarios is open.

%% file: appendix/main.tex
\input{appendix/1_matching_pursuit}
\input{appendix/2_robustness}
\input{appendix/3_crosspair}
\input{appendix/4_rlens}
\input{appendix/5_hyperparameters}
\input{appendix/6_example}

%% file: appendix/1_matching_pursuit.tex
\section{Matching pursuit for J-space extraction}
\label{app:matching_pursuit}

The J-space component of the residual activation $h$ is recovered by matching pursuit against the J-lens dictionary $\{v_w\}_{w \in V}$, where $v_w$ is the J-lens direction for token $w$. Starting from residual $\rho_0 = h$, at each step $k$ we select the top-scoring token
\begin{equation}
w_k = \arg\max_w v_w^\top \rho_{k-1},
\end{equation}
refit the active coefficients by nonnegative least squares,
\begin{equation}
\mathbf{a} = \arg\min_{\mathbf{a} \geq 0} \|h - V_k \mathbf{a}\|^2 \quad \text{where } V_k = [v_{w_1}, \dots, v_{w_k}],
\end{equation}
and update the residual $\rho_k = h - V_k \mathbf{a}$. We stop at the sparsity cap ($K = 16$ for probing, $K = 10$ for ablation) or when $\max_w v_w^\top \rho_{k-1} \leq 0$. The J-space component is $V_K \mathbf{a}$.

%% file: appendix/2_robustness.tex
\section{Robustness of the decodability result}
\label{app:robustness}

The J-space null of \Cref{sec:results:decodability} is not an artifact of probe class, regularization, seed, or layer choice.

\paragraph{Top J-lens atoms.} The replication gate passes on all three models. The top J-lens atoms mix tokenizer noise with grammatical vocabulary: \textit{sentence}, \textit{noun}, \textit{Passive}, and \textit{pronoun} on Gemma-3-27B-IT, \textit{verb} and \textit{ambiguous} on Qwen3.6-27B.

\paragraph{Nonlinear probes.} A quadratic probe on J-space, using ridge regression on the concatenation of the standardized activation with random pairwise products $x_i \cdot x_j$, reaches 0.522 at the entity token on Qwen3.6-27B and 0.449 on Gemma-3-12B. A random-Fourier-features probe on random ReLU features reaches similar values. Under each probe family, the matched random subspace scores higher than J-space.

\paragraph{Ridge penalty sweep.} Sweeping the ridge penalty across seven orders of magnitude lifts J-space to 0.645 at best, while the random subspace lifts to 0.678 to 0.724 across the same sweep. Nested cross-validation, in which the penalty is chosen on inner folds within the training set and never on the held-out pair, preserves the ordering.

\paragraph{Seed and layer stability.} J-space accuracies are deterministic at a fixed layer, so seed variation only affects the random subspace, which moves between 0.313 and 0.330 across five seeds on Gemma-3-12B. Repeating the probe at layers 18, 36, and 44 of Gemma-3-12B and at layers 24, 36, 48, and 59 of Qwen3.6-27B, J-space accuracy stays in a narrow band. At layer 18 of Gemma-3-12B, J-space and the random subspace tie at 0.323 and 0.322.

%% file: appendix/3_crosspair.tex
\section{Cross-pair transfer at the sentence-final position}
\label{app:crosspair}

If a probe has learned a genuine role axis, leave-one-pair-out cross-pair transfer should hold: a probe trained on \textit{doctor}/\textit{nurse} and \textit{teacher}/\textit{student} should classify held-out \textit{driver}/\textit{passenger} stimuli above chance. It does not, on any activation source, at the sentence-final position.

Cross-pair transfer at the sentence-final position lands at 0.320 on all three models for J-space, at 0.283 for the full residual, and at 0.281 for the orthogonal remainder. Every source inverts below chance, including the one that decodes role at 0.989 within pair. The pattern appears under R-lens on Gemma-3-27B-IT as well.

The entity token behaves differently. Table~\ref{tab:crosspair_entity} reports leave-one-pair-out held-out accuracy at the entity token for all four activation sources on all three models. The full residual stream and the orthogonal remainder transfer above chance (0.594 to 0.644 and 0.603 to 0.626); the matched-rank random subspace transfers at 0.575 to 0.713; J-space stays at chance (0.510 to 0.524). The anti-transfer is therefore a sentence-final-position phenomenon, and at the entity token the full residual and orthogonal remainder do carry a filler-general role axis while J-space does not.

\begin{table}[htb]
\centering
\captionsetup{labelfont=bf,textfont=it}
\caption{Leave-one-pair-out cross-pair transfer at the entity token. Chance is 0.50.}
\label{tab:crosspair_entity}
\small
\begin{tabular}{lccc}
\toprule
Decoded from & Qwen3.6-27B & Gemma-3-12B & Gemma-3-27B-IT \\
\midrule
Full residual stream            & 0.594 & 0.644 & 0.601 \\
Orthogonal remainder            & 0.626 & 0.603 & 0.605 \\
Random subspace (matched rank)  & 0.713 & 0.613 & 0.575 \\
J-space                         & 0.524 & 0.514 & 0.510 \\
\bottomrule
\end{tabular}
\end{table}

The uniformity of the inversion across sources (including the full residual) at the sentence-final position means that pattern reflects the probing setup rather than a property of J-space. By contrast, the clean separation at the entity token, where every source except J-space transfers above chance, shows that the within-pair J-space result reflects a property of J-space itself, not of the probing setup. We report within-pair rather than cross-pair as the primary test for this reason.

%% file: appendix/4_rlens.tex
\section{Cross-lens replication of the Gemma-3-27B-IT ablation}
\label{app:rlens}

The binding deficit on Gemma-3-27B-IT at the entity token, reported in \Cref{sec:results:ablation}, replicates across three lens variants:

\begin{itemize}
    \item J-lens from the primary release (\texttt{neuronpedia/jacobian-lens}): 0.284 (0.275 binding-specific).
    \item Matched J-lens from the R-lens release (\cite{Blank2026}): 0.334 (0.299 binding-specific).
    \item R-lens itself (\cite{Blank2026}): 0.297 (0.280 binding-specific).
\end{itemize}

The three deficits are all well above the pre-registered 0.05 causal-involvement floor and all exceed their respective matched random-subspace controls. The ablation asymmetry is therefore a property of the workspace-lens paradigm on Gemma-3-27B-IT rather than of the J-lens variant used in the primary run.

%% file: appendix/5_hyperparameters.tex
\section{Hyperparameters, calibration, and compute}
\label{app:hyperparameters}

This appendix collects the numerical hyperparameters, calibrated values, and compute details for the runs in \Cref{sec:experimental_setup}.

\paragraph{Probes and direction fitting.} The ridge penalty robustness sweep spans seven orders of magnitude ($10^{-4}$ to $100$). Nonlinear probes use 2048 random features (quadratic and random-Fourier variants). Bootstrap direction stability uses 200 resamples with a threshold of 0.8 mean cosine similarity; directions below this threshold are not used.

\paragraph{Layer indices.} The primary read layer is L48 on Qwen3.6-27B (of 64), L36 on Gemma-3-12B (of 48), and L46 on Gemma-3-27B-IT (of 62). The Gemma-3-27B-IT choice lands on a rounding boundary ($75\% \times 62 = 46.5$); we used L46. For the layer sweep on Gemma-3-27B-IT, the workspace band edges are L24 and L57, matching the source paper's reindexed L38 and L92 \citep{gurnee2026workspace}.

\paragraph{Push coefficients.} Calibrated entity-token coefficients: Qwen3.6-27B with difference-of-means $c = 8$; Qwen3.6-27B with LRE $c = 4$; Gemma-3-12B with LRE $c = 256$; Gemma-3-27B-IT with LRE $c = 256$. The Gemma-3-27B-IT dose-response also ran $c = 64$ and $c = 128$ as separate runs.

\paragraph{Ablation.} Ablation uses $K_{\text{ablate}} = 10$ top J-lens vectors, selected per stimulus by matching pursuit. The random-subspace control basis is seeded by family index and drawn fresh per family. The binding-specific deficit floor for a causal-involvement claim is 0.05 (as a share of the no-edit binding margin). The difficulty-matching tolerance between binding and recall baseline accuracy is 0.02.

\paragraph{Statistical tests.} Both bootstrap and permutation use 10,000 resamples at $\alpha = 0.05$ with 95\% confidence intervals.

\paragraph{Verdict classifier thresholds.} A meaningful effect needs Cohen's $d \ge 0.5$. An informative null needs a $d$-CI width $< 1.0$. Positive binding requires the effect to hold in at least two of the three agent/patient constructions (active-passive, cleft, relative clause; dative measures giver vs.\ recipient and is reported separately). The null band is the 95th-percentile band of the control scores.

\paragraph{Seeds.} Three of the five seeds are complete on Qwen3.6-27B (0, 42, 7), with two (123 and 2024) outstanding. Both Gemma models have all five seeds.

\paragraph{Compute.} Experiments ran on 80\,GB A100 or H100 nodes at bfloat16, and the 27B models were sharded across two 40\,GB cards. We did not use 4-bit quantization because the LRE gradient estimator requires backward passes.

%% file: appendix/6_example.tex
\section{A worked stimulus family}
\label{app:worked-example}

Table~\ref{tab:worked-example} follows a single stimulus family through the readout, so
that the measurement the rest of the paper builds on can be checked against
concrete numbers. The family holds one target entity fixed, the \emph{doctor},
and crosses the role that entity fills with its linear position in the
sentence, resulting in the four cells shown. The second participant, the
\emph{lawyer}, changes position alongside the target but is never the entity
being scored for role, and the verb is held fixed across the four cells so that
nothing but the syntactic frame varies.

\begingroup
\centering
\captionsetup{labelfont=bf,textfont=it}
\captionof{table}{The four cells of one stimulus family. All four are read with the same probe and scored from the same two vocabulary entries, with
$\ell = \log\big(P(\text{doctor})/P(\text{lawyer})\big)$ positive exactly when
the target is the agent and negative when it is the patient.}
\label{tab:worked-example}
\vspace{6pt}
{\small \textbf{Probe (all four cells)}: \emph{Question: Who treated someone? Answer: The\_}}
\vspace{4pt}
\small
\begin{tabular}{lllrrr}
\toprule
\textbf{Role} & \textbf{Position} & \textbf{Stimulus sentence} &
  $P(\text{doctor})$ & $P(\text{lawyer})$ & $\ell$ \\
\midrule
agent   & first  & The doctor treated the lawyer.        & 0.99376 & 0.00016 & $+8.75$ \\
agent   & second & The lawyer was treated by the doctor. & 0.99543 & 0.00023 & $+8.37$ \\
patient & first  & The doctor was treated by the lawyer. & 0.26712 & 0.72611 & $-1.00$ \\
patient & second & The lawyer treated the doctor.        & 0.08452 & 0.90863 & $-2.38$ \\
\bottomrule
\end{tabular}
\par
\endgroup

Every cell is read with the same probe and scored from the same two vocabulary
entries, which is what makes the four readouts comparable. We append
\emph{Question: Who treated someone? Answer: The} to the stimulus, take the
next-token distribution over the full vocabulary, and keep only
$P(\text{doctor})$ and $P(\text{lawyer})$, discarding the remaining probability
mass. The readout is the log-odds between those two, $\ell$, which when positive indicates that the model names the target as the agent,
and when negative indicates that it names the other party instead.

The sign structure in the table is what Equation~2 uses. Both agent cells
come out strongly positive, at $+8.75$ and $+8.37$, and both patient cells
negative, at $-1.00$ and $-2.38$, showing that the sign tracks role.
Position moves the magnitude within each sign without changing it, which is why
averaging over position inside a family removes word-order effects while
leaving the role contrast intact. The asymmetry in magnitude between the agent
and patient cells also shows why the binding score is a difference of
differences rather than a raw log-odds; the model is far more confident in naming
the agent than in declining to name the patient, and only a paired contrast
removes that baseline.